\pdfoutput=1

\documentclass[11pt]{article}

\usepackage[final]{acl}
\usepackage{natbib}
 \usepackage{amsmath} 
\usepackage{amssymb}
\usepackage{times}
\usepackage{latexsym}
\usepackage{booktabs}
\usepackage{multirow}
\usepackage[T1]{fontenc}

\usepackage[utf8]{inputenc}

\usepackage{microtype}

\usepackage{inconsolata}
\usepackage{hyperref}
\usepackage{graphicx}
\usepackage{amssymb}
\usepackage{bbding}

\title{TrInk: Ink Generation with Transformer Network}

\author{
  \textbf{Zezhong Jin}\textsuperscript{1}\thanks{\ \ This work was conducted during an internship at Microsoft Research Asia.},
  \textbf{Shubhang Desai}\textsuperscript{2},
  \textbf{Xu Chen}\textsuperscript{2},
  \textbf{Biyi Fang}\textsuperscript{2},
  \textbf{Zhuoyi Huang}\textsuperscript{2},
  \textbf{Zhe Li}\textsuperscript{1}, \\
  \textbf{Chong-Xin Gan}\textsuperscript{1},
  \textbf{Xiao Tu}\textsuperscript{2},
  \textbf{Man-Wai Mak}\textsuperscript{1}\thanks{Corresponding author.},
  \textbf{Yan Lu}\textsuperscript{2},
  \textbf{Shujie Liu}\textsuperscript{2}\footnotemark[2] \\
  \textsuperscript{1}The Hong Kong Polytechnic University \\
  \textsuperscript{2}Microsoft Corporation \\
}

\begin{document}
\maketitle
\begin{abstract}
In this paper, we propose TrInk, a Transformer-based model for ink generation, which effectively captures global dependencies. To better facilitate the alignment between the input text and generated stroke points, we introduce scaled positional embeddings and a Gaussian memory mask in the cross-attention module. Additionally, we design both subjective and objective evaluation pipelines to comprehensively assess the legibility and style consistency of the generated handwriting. Experiments demonstrate that our Transformer-based model achieves a 35.56\% reduction in character error rate (CER) and an 29.66\% reduction in word error rate (WER) on the IAM-OnDB dataset compared to previous methods. We provide an demo page with handwriting samples from TrInk and baseline models at: \url{https://akahello-a11y.github.io/trink-demo/}
\end{abstract}
  
\section{Introduction}
\label{sec: intro}
Handwriting synthesis is the task of automatically generating realistic handwritten text from digital inputs. Automatic handwritten text generation can support a wide range of applications, including digital note-taking, educational tools, and generating training data to improve optical character recognition (OCR) systems \cite{li2023trocr, fujitake2024dtrocr, yeleussinov2023improving}. However, due to the complex temporal dynamics and variability inherent in human handwriting, generating high-quality handwritten samples still faces challenges. 

Deep learning-based handwritten text generation approaches can be roughly divided into image-based offline and stroke-based online methods, the latter also referred to as ink generation. Offline handwriting synthesis focuses on producing a static image of handwritten text \cite{chang2018generating, alonso2019adversarial, kang2020ganwriting, haines2016my,bhunia2021handwriting}. In contrast, online handwriting synthesis (also called ink generation) aims to generate a time-ordered sequence of pen-tip coordinates along with pen-state indicators (e.g., pen-up and pen-down), thereby reconstructing the full dynamic trajectory of the writing process. 
Compared with offline approaches, online handwriting synthesis (ink generation) outputs lightweight stroke vectors that can be rendered at any resolution, making them easy to transmit and display consistently across diverse devices. In this work, we focus on  ink generation to generate handwriting samples that are stylistically consistent and highly legible.


Recent research on ink generation has predominantly relied on sequential models \cite{graves2013generating, aksan2018deepwriting, chang2022style}. \citet{graves2013generating} leverages an LSTM-based network to predict future stroke points from the current pen position based on the given text. \citet{aksan2018deepwriting} introduces a conditional variational RNN that improves the model’s capacity to capture handwritten digits. Building on \citet{graves2013generating}, \citet{chang2022style} introduces style equalization method, equipped with a style encoder to explicitly model the style information. 

While these approaches have demonstrated promising results, they remain fundamentally constrained by the sequential nature of recurrent architectures, which limits their ability to model long-range dependencies and hinders parallel training. Furthermore, alignment between the input text and generated strokes often requires careful design, such as attention windowing. Motivated by the success of Transformer \cite{vaswani2017attention} in various generative tasks \cite{li2019neural, chen2020multispeech, ding2021cogview, chang2023muse, ma2024latte}, we propose TrInk (Transformer for Ink Generation), a fully attention-based model tailored for ink generation. The encoder ingests the target text sequence, through multi-head self-attention, yields a contextual content representation for every character. The decoder receives the character representations together with the previous generated stroke points, and applies multi-head self- and cross-attention to compute decoder hidden states. These decoder hidden states are fed into a mixture-density network, which outputs a Gaussian mixture distribution from which the next pen offset and pen state are sampled. To improve alignment between the text and the stroke sequence, we apply a Gaussian memory mask to the cross-attention matrix, constraining the decoder’s focus to progress strictly left-to-right along the encoded text as strokes are generated. We apply a learnable scale to sinusoidal positional embeddings to handle differences between text and ink points. Our main contributions are summarized as follows:
\begin{enumerate}
    \item To the best of our knowledge, TrInk is the first to employ a Transformer encoder–decoder architecture for ink generation.
    \item TrInk introduces a Gaussian memory mask to ensure the generated ink points follows the natural writing order, and a scale factor for the position embeddings to model the differing charateristics of the text and the ink points.
    \item Our experimental results show that our proposed TrInk yields substantially higher legibility, particularly on long text, than the previous.

\end{enumerate}

\section{Method}
\label{sec: method}

\begin{figure*}[!t]
    \centering
    \includegraphics[width=\linewidth]{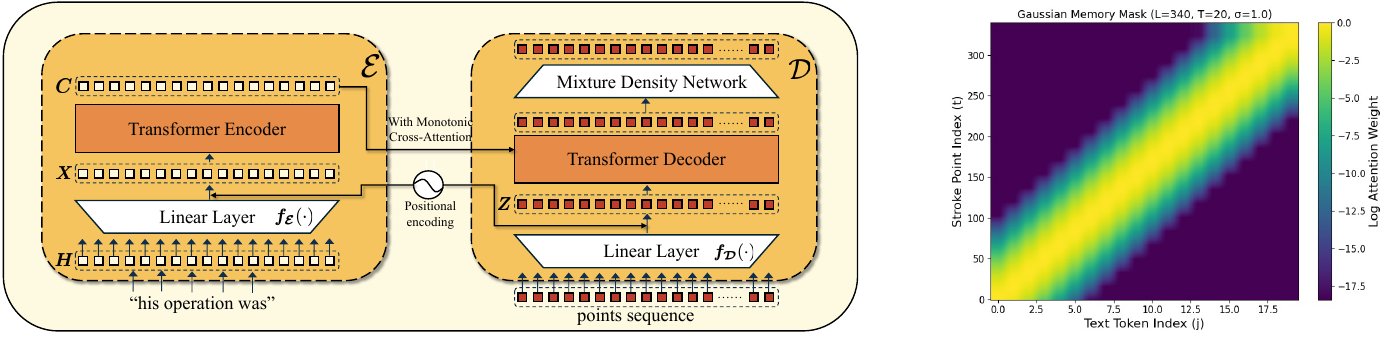}
    \caption{The proposed TrInk framework (left) and Gaussian memory mask (right).}
    \label{fig:framework}
\end{figure*}

The Framework of TrInk comprises two main components: an encoder $\cal E$ and a decoder $\cal D$. Encoder $\cal E$ aims to map the input text into a sequence of context-aware distributed representations, with each vector encoding a token in relation to its surrounding context. Decoder $\cal D$ aims to take the encoder’s content vectors along with the previous stroke points and, at each time step, predict Gaussian distributions for the next pen coordinate and the stroke-end probability.

\subsection{Encoder}
From Fig.~\ref{fig:framework}, we transform each character of the input "his operation was" (including the blank spaces) into a one-hot vector, yielding $\boldsymbol{H} =[\boldsymbol{h}_1,\boldsymbol{h}_2,\dots,\boldsymbol{h}_T]$, and $\boldsymbol{h}_i \in \mathbb{R}^{|V|}$, where $|V|$ denotes the vocabulary size and $T$ denotes the number of tokens in the text. After the linear projection and positional encoding, we obtain the Transformer encoder input $\boldsymbol{X} = [\boldsymbol{x}_1,\boldsymbol{x}_2,\dots,\boldsymbol{x}_T],\quad \boldsymbol{x}_i \in \mathbb{R}^d,$ where $d$ is the hidden‐state dimension of the Transformer encoder. The high-dimensional text representation $\boldsymbol{C}$ generated by the encoder is then fed into the Transformer decoder to serve as the memory for cross-attention.

\subsection{Scaled Positional Encoding}
To account for the sequential order of both text tokens and stroke points in ink generation, we inject absolute position information using sinusoidal positional embeddings, as defined below:
\begin{equation}
\small
\begin{aligned}
PE(pos,2i)   &= \sin\!\Bigl(\frac{pos}{10000^{\frac{2i}{d_{}}}}\Bigr),\\
PE(pos,2i+1) &= \cos\!\Bigl(\frac{pos}{10000^{\frac{2i}{d_{}}}}\Bigr),
\end{aligned}
\end{equation}
where $pos$ denotes the position index and $d$ is the model's hidden dimension. Because the encoder’s domain is text and the decoder’s domain is stroke points, using fixed positional embeddings alone cannot properly capture the differing scales and characteristics of these two inputs. We therefore employ these sinusoidal positional embeddings with trainable weights so that the embeddings can adaptively fit the output scales of both the encoder’s and the decoder’s linear layers, following \cite{li2019neural}, as shown in Eq.~\ref{eq: positional}
\begin{equation}
\small
    \boldsymbol{x}_i = \boldsymbol{f_{\cal E}}(\boldsymbol{h}_i) + \alpha PE(i)
    \label{eq: positional}
\end{equation}
where $\alpha$ is the trainable weight. A similar formulation with a separate scaling parameter is applied in the decoder.
\subsection{Decoder with Monotonic Cross-Attention}
Each stroke point is represented as a 3-dimensional vector $[\Delta x, \Delta y, s],$ where $\Delta x$ and $\Delta y$ are the offsets along the $x$- and $y$-axes, and $s \in \{0,1\}$ denotes the pen state (0 = pen-down, 1 = pen-up). After the linear projection and positional encoding, we obtain $\boldsymbol{Z} =[\,\boldsymbol{z}_1,\boldsymbol{z}_2\dots,\boldsymbol{z}_L\,]$, where $\boldsymbol{z}_i \in \mathbb{R}^d$ and $L$ is the number of stroke points.

Given the stroke embedding sequence $\boldsymbol{Z}$, the Transformer decoder first applies masked self-attention to enforce autoregressive dependencies among stroke points. It then performs cross-attention with the text representations $\boldsymbol{C}$ to align each generated stroke with the corresponding text content. To ensure that each decoding step attends to the most relevant region of the input text, we introduce a Gaussian-shaped cross-attention mask. For each decoder time step $t \in [1, L]$, we define its corresponding attention center $\mu_t$ on the text sequence $\boldsymbol{C}$ as:
\begin{equation}
\small
    \mu_t = \min\left( \frac{t}{r},\ T - 1 \right)
    \label{eq: center}.
\end{equation}
 $r$ denotes the average number of stroke points per character, estimated from the training data. Gaussian function centered at $\mu_t$ is used for each decoder step $t$, ensuring higher attention weights for text positions near the center and lower weights for distant ones. For each decoder step $t$ and encoder position $j \in [1, T]$, the attention weight is defined as:
\begin{equation}
\small
     \boldsymbol{A}_{t,j} = \exp\left( - \frac{(j - {\mu}_t)^2}{2{\sigma}^2} \right),
     \label{eq: weight}
\end{equation}
where $\sigma$ is a controllable parameter that determines the sharpness of the Gaussian distribution. We apply the logarithm to the $\boldsymbol{A}_{t,j}$ to obtain the cross-attention mask $\boldsymbol{M}_{t,j} = \log(\boldsymbol{A}_{t,j})$, allowing it to be added directly to the attention logits before the softmax, similar to standard attention masking in Transformers. This log-space formulation helps maintain numerical stability by avoiding extremely small values in the Gaussian tails. Gaussian cross-attention mask ensures that attention shifts monotonically from left to right across the input text. At each decoding step, encoder positions $j$ near the center $\mu_t$ receive higher attention scores, while positions farther from $\mu_t$ are gradually suppressed, as illustrated in the right side of Fig.~\ref{fig:framework}.

After the Transformer decoder, we adopt a mixture density network (MDN) \cite{bishop1994mixture} to model the output distribution, following the strategy of \cite{graves2013generating}. Instead of directly producing continuous stroke points, the decoder outputs a $(6K + 2)$-dimensional vector at each time step, where $K$ is the number of Gaussian mixture components. This vector encodes the parameters of $K$ bivariate Gaussian distributions—mixture weights, means, standard deviations, and correlation coefficients—together with two additional scalars: an end-of-stroke probability and a sequence-level stop probability. During inference, the actual pen-point coordinates are sampled from the predicted mixture of Gaussians, rather than being deterministically generated.

We adopt the same training objective as \citet{graves2013generating}, minimizing the negative log-likelihood of the ground-truth trajectory. The loss function comprises three components: a mixture density loss for predicting stroke offsets, a Bernoulli loss for the end-of-stroke indicator, and a Bernoulli loss for determining sequence termination.
\vspace{-2mm}
\begin{table*}[!t]
\small
\centering
\begin{tabular}{l|cc|cc|c}
\toprule
\multirow{3}{*}{\textbf{Method}} & \multicolumn{5}{c}{\textbf{IAM-OnDB}}\\
\cline{2-6}
& \multicolumn{2}{c|}{\textbf{Full}} & \multicolumn{2}{c|}{\textbf{Long Texts}} & \multicolumn{1}{c}{\textbf{Short Texts}} \\
\cline{2-6}
 & \textbf{CER, \% }$\downarrow$ & \textbf{WER, \%} $\downarrow$ &\textbf{CER, \% }$\downarrow$ & \textbf{WER, \%} $\downarrow$ & \textbf{CER, \% }$\downarrow$  \\ 
\midrule
AlexRNN \cite{graves2013generating}  & 9.0 & 53.6 & 15.6& 48.6 & 27.8  \\
AlexRNN (Top-k)&8.8 &42.6 & 10.0 & 40.1 & 18.3\\
Style Equalization \cite{chang2022style}&8.7& 47.4 &11.6 & 46.0& 24.4\\
Style Equalization (Top-k)&6.5 &40.0 &8.7 &39.7 &18.2 \\
TrInk & 8.5 & 43.2  & 9.3 & 43.3 & 21.9  \\
TrInk (Top-k) & \textbf{5.8} & \textbf{37.7} & \textbf{6.8} &\textbf{36.3} & \textbf{17.6} \\
\bottomrule
\end{tabular}
\caption{Comparison of different methods on three test sets ($\mathrm{k}=5$).}
\label{tab:main}
\end{table*}

\section{Experiments}
\label{sec: experiment}
\subsection{Datasets}
\label{subsec: dataset}
The original training dataset was collected from over 5,000 writers and was initially used for online handwriting recognition tasks. However, we observed that some ink samples were overly cursive that benefit the training of robust recognition systems but are not suitable for generating realistic handwriting. To tackle this issue, we leveraged an OCR engine to filter the dataset, selecting a curated subset of 600,000 high-quality ink samples, optimally prepared for handwriting generation. For evaluation, we use the IAM-OnDB \cite{liwicki2005iam} test set, which is the most popular dataset for handwritten text recognition. We divide the test set into three subsets: full, short, and long. The long subset comprises samples exceeding 40 characters, while the short subset includes those with fewer than 10 characters.

\subsection{Evaluation Pipeline}
Inspired by text-to-speech evaluation protocols, we divide our evaluation into subjective and objective assessments. For the subjective evaluation, human raters fluent in English scored the generated handwriting samples based on legibility and stylistic consistency, each on a 1–5 scale, with higher scores indicating better quality. For the objective evaluation, we utilize a state-of-the-art OCR model \cite{li2023trocr} to recognize the generated samples, comparing the outputs to the ground-truth text to compute the Character Error Rate (CER) and Word Error Rate (WER) as quantitative measures of legibility. Since our focus is on online handwriting synthesis (ink generation), which generates stroke-point sequences character by character while capturing the dynamics of writing, the evaluation protocol is naturally different from that of offline, image-based handwriting generation. The two tasks differ in their data representation (stroke trajectories versus images), model objectives (temporal and spatial consistency versus visual realism), and evaluation protocols (sequence-oriented metrics such as CER and WER versus image-oriented metrics such as FID and SSIM). Because of these fundamental differences, direct comparisons with offline methods are not meaningful. Accordingly, we deliberately compare against AlexRNN and Style Equalization, which are both online methods, to ensure fairness and relevance.
\subsection{Main Results}
Table~\ref{tab:main} presents the objective evaluation results on the IAM-OnDB test set. All models were trained on the same dataset, as described in Section~\ref{subsec: dataset}. Our pipeline employs a Top-$k$ sampling strategy where $k$ candidate handwriting samples are first generated, then ranked by TrOCR according to their CER scores against the ground-truth text, with the optimal sample (minimum CER) selected as final output. As shown in Table~\ref{tab:main}, TrInk consistently outperforms all baselines, including both the standard AlexRNN and its variant with style equalization, across all evaluation settings. TrInk with the Top-$k$ strategy achieves the best performance, with a 35.56\% reduction in CER and a 29.66\% reduction in WER on the full test set compared to AlexRNN. For long-text generation, TrInk shows even greater improvements, with a 56.41\% reduction in CER and a 25.31\% reduction in WER compared to AlexRNN. These reductions further highlight the effectiveness of TrInk.

\begin{figure}[!t]
    \centering
    \includegraphics[width=0.7\linewidth]{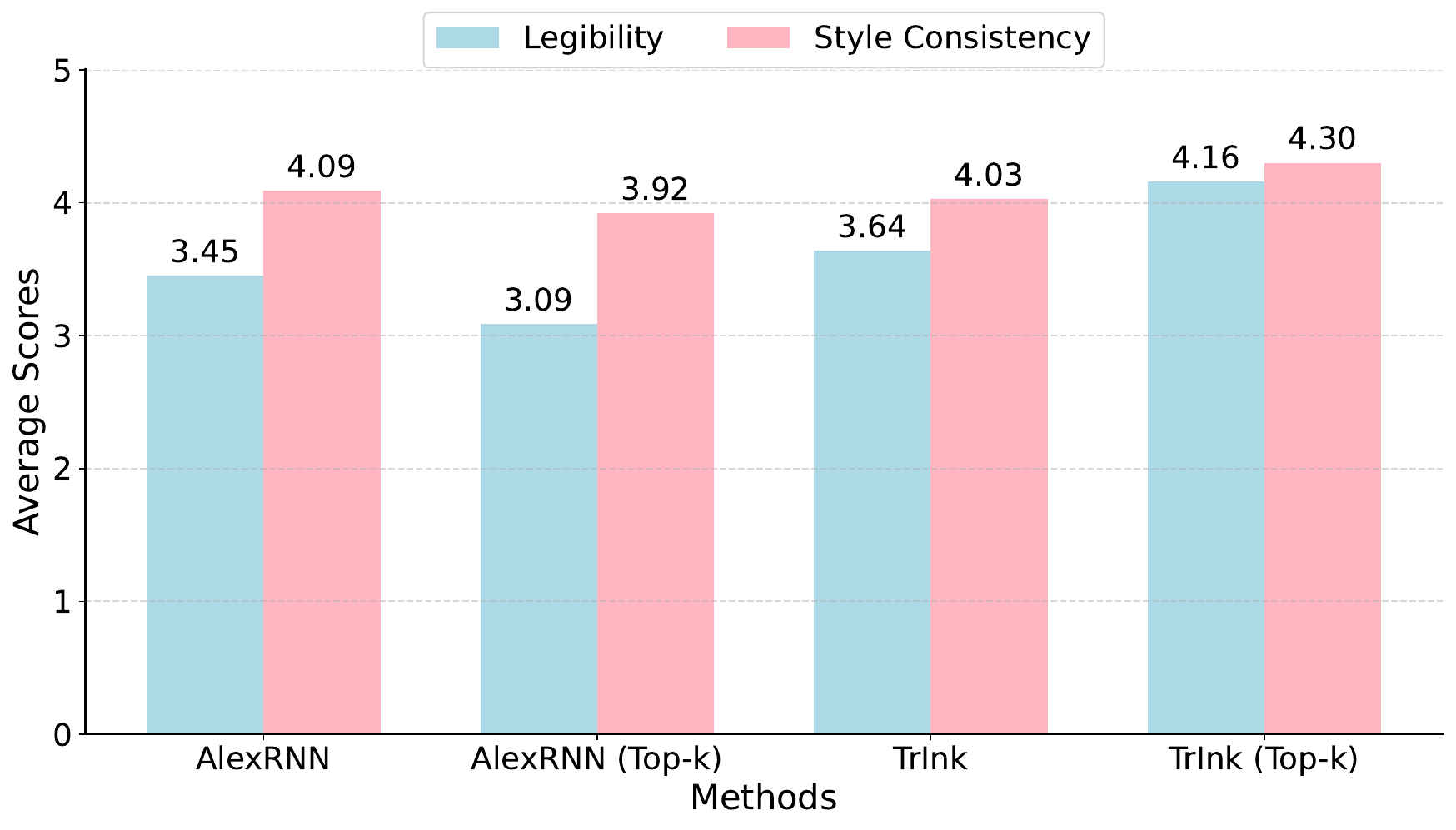}
    \caption{Subjective Evaluation of Handwriting Quality Across Methods.}
    \label{fig: subjective}
\end{figure}

Figure.~\ref{fig: subjective} presents the results of our subjective experiments. The final scores for each method were the average ratings for two metrics: style consistency and legibility. From Figure.~\ref{fig: subjective}, we can observe that TrInk outperforms AlexRNN in both metrics, further validating the effectiveness of TrInk.

\subsection{Ablation Study}
\begin{figure}[!t]
    \centering
    \includegraphics[width=0.7\linewidth]{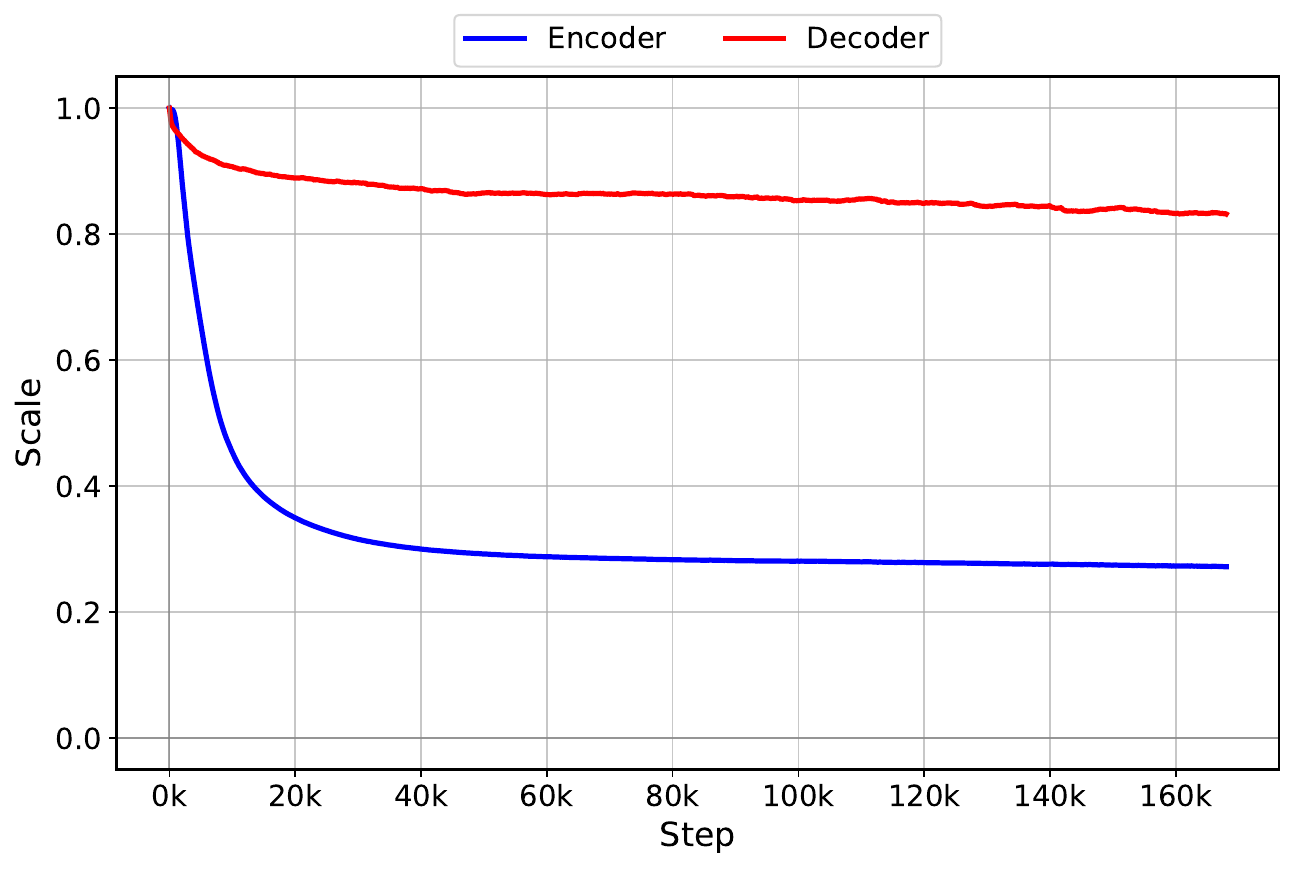}
    \caption{Trainable Positional Encoding Weights (Encoder and Decoder) Over Training.}
    \label{fig:scale pe}
\end{figure}
\begin{table}[!t]
    \small
    \centering
    \captionsetup{font=small}
    \caption{Effectiveness of Gaussian Memory Mask in Cross-Attention Alignment.}
    \resizebox{\linewidth}{!}{
    \begin{tabular}{c|c|c}
    \toprule[1pt]
     \bf{Method}& \bf{Gaussian memory mask} &\bf{CER, \% $\downarrow$} \\  
     \toprule[0.5pt]
     \midrule[0.5pt]
     \multirow{2}{*}{TrInk} & \XSolidBrush& 70.1\\
     & \Checkmark  & 8.5 \\
     \bottomrule[1pt]
    \end{tabular}
    }
    \label{tab:gaussian memory mask}
\end{table}

\begin{table}[!t]
    \small
    \centering
    \captionsetup{font=small}
    \caption{Effect of different memory mask functions on TrInk performance (IAM-OnDB Test Set).}
    \resizebox{0.75\linewidth}{!}{
    \begin{tabular}{c|c|c}
    \toprule[1pt]
     \bf{Method} & \bf{Mask Function} & \bf{CER, \% $\downarrow$} \\  
    \toprule[0.5pt]
    \midrule[0.5pt]
     \multirow{3}{*}{TrInk} & Uniform     & 14.9 \\
                            & Exponential & 14.1 \\
                            & Gaussian    & \bf 8.5 \\
    \bottomrule[1pt]
    \end{tabular}
    }
    \label{tab:mask}
\end{table}

\begin{table}[!t]
    \small
    \centering
    \captionsetup{font=small}
    \caption{Effectiveness of $\sigma$ on Model performance.}
    \resizebox{0.65\linewidth}{!}{
    \begin{tabular}{c|c|c}
    \toprule[1pt]
     \bf{Method}& \bf{$\sigma$ Value} &\bf{CER, \% $\downarrow$} \\  
     \toprule[0.5pt]
     \midrule[0.5pt]
     \multirow{3}{*}{TrInk} & 2.0& 8.7\\
     & 1.0  & 8.5 \\
     &0.5& \bf 8.5 \\
     \bottomrule[1pt]
    \end{tabular}
    }
    \label{tab:sigma}
\end{table}
Figure.~\ref{fig:scale pe} illustrates the changes of the two trainable weights in the positional encoding for the encoder and decoder during training. Notably, these weights converge to different values, indicating a significant discrepancy, showing that adopting fixed positional embeddings may fail to capture the differing scales and characteristics of the two input modalities. 

To examine the role of the Gaussian function in our memory mask design, we conducted ablation experiments by replacing the Gaussian kernel with alternative formulations. Specifically, the uniform mask assigns equal weights to all memory positions within a fixed local window, while the exponential mask applies a sharper decay around the center position with weights defined by an exponential function. As shown in Table~\ref{tab:mask}, the Gaussian mask achieves the lowest Character Error Rate (CER) of 8.5\% on the IAM-OnDB test set, whereas the uniform and exponential variants yield significantly higher CERs of 14.9\% and 14.1\%, respectively. These results demonstrate that the Gaussian formulation provides smoother and more robust alignment between stroke points and text tokens, while the uniform and exponential alternatives either oversimplify the weighting scheme or introduce overly sharp decays, both of which lead to degraded performance.

We also investigated the effect of the Gaussian memory mask on the IAM-OnDB full test set, as shown in Table~\ref{tab:gaussian memory mask}. Removing the Gaussian memory mask leads to a significant drop in the legibility of the generated samples. This is mainly because the model fails to learn proper alignment between the text and stroke points without the guidance of the mask. We also investigate the impact of the hyperparameter $\sigma$ in the Gaussian memory mask. As shown in Table~\ref{tab:sigma}, our model is relatively insensitive to variations in $\sigma$, with the best performance observed when $\sigma = 1.0$ and $\sigma = 0.5$.

\vspace{-2mm}
\section{Conclusion}
\label{sec:conclusion}
This paper presents TrInk, the first ink-generation model built on a Transformer encoder–decoder architecture. To achieve precise alignment between input text and generated stroke sequences, we introduce a scaled positional encoding with learnable weights and a Gaussian memory mask. We also devise both subjective and objective evaluation protocols for ink generation. Experimental results demonstrate that TrInk markedly outperforms traditional LSTM-based approaches, producing handwriting samples with superior style consistency and legibility.

\section{Limitations}
\label{sec: limitations}
Despite the promising results, TrInk has two limitations. First, training our Transformer-based architecture requires considerable computational resources. The increased model capacity and parallel attention mechanisms lead to higher memory consumption and longer convergence time compared to lightweight RNN-based alternatives.

Second, our current experiments are conducted solely on English handwriting datasets. As handwriting conventions vary significantly across scripts and languages (e.g., cursive structures in Arabic, character-based layouts in Chinese), it remains unclear how well our model generalizes to multilingual settings. Developing a unified ink generation framework capable of generating stylistically consistent samples across multiple languages would be an important direction for future work.
\bibliography{custom}

\begin{thebibliography}{19}
\providecommand{\natexlab}[1]{#1}

\bibitem[{Aksan et~al.(2018)Aksan, Pece, and Hilliges}]{aksan2018deepwriting}
Emre Aksan, Fabrizio Pece, and Otmar Hilliges. 2018.
\newblock Deepwriting: Making digital ink editable via deep generative modeling.
\newblock In \emph{Proceedings of the 2018 CHI conference on human factors in computing systems}, pages 1--14.

\bibitem[{Alonso et~al.(2019)Alonso, Moysset, and Messina}]{alonso2019adversarial}
Eloi Alonso, Bastien Moysset, and Ronaldo Messina. 2019.
\newblock Adversarial generation of handwritten text images conditioned on sequences.
\newblock In \emph{2019 international conference on document analysis and recognition (ICDAR)}, pages 481--486. IEEE.

\bibitem[{Bhunia et~al.(2021)Bhunia, Khan, Cholakkal, Anwer, Khan, and Shah}]{bhunia2021handwriting}
Ankan~Kumar Bhunia, Salman Khan, Hisham Cholakkal, Rao~Muhammad Anwer, Fahad~Shahbaz Khan, and Mubarak Shah. 2021.
\newblock Handwriting transformers.
\newblock In \emph{Proceedings of the IEEE/CVF international conference on computer vision}, pages 1086--1094.

\bibitem[{Bishop(1994)}]{bishop1994mixture}
Christopher~M Bishop. 1994.
\newblock Mixture density networks.

\bibitem[{Chang et~al.(2018)Chang, Zhang, Pan, and Meng}]{chang2018generating}
Bo~Chang, Qiong Zhang, Shenyi Pan, and Lili Meng. 2018.
\newblock Generating handwritten chinese characters using cyclegan.
\newblock In \emph{2018 IEEE winter conference on applications of computer vision (WACV)}, pages 199--207. IEEE.

\bibitem[{Chang et~al.(2023)Chang, Zhang, Barber, Maschinot, Lezama, Jiang, Yang, Murphy, Freeman, Rubinstein et~al.}]{chang2023muse}
Huiwen Chang, Han Zhang, Jarred Barber, AJ~Maschinot, Jose Lezama, Lu~Jiang, Ming-Hsuan Yang, Kevin Murphy, William~T Freeman, Michael Rubinstein, and 1 others. 2023.
\newblock Muse: Text-to-image generation via masked generative transformers.
\newblock \emph{arXiv preprint arXiv:2301.00704}.

\bibitem[{Chang et~al.(2022)Chang, Shrivastava, Koppula, Zhang, and Tuzel}]{chang2022style}
Jen-Hao~Rick Chang, Ashish Shrivastava, Hema Koppula, Xiaoshuai Zhang, and Oncel Tuzel. 2022.
\newblock Style equalization: Unsupervised learning of controllable generative sequence models.
\newblock In \emph{International Conference on Machine Learning}, pages 2917--2937. PMLR.

\bibitem[{Chen et~al.(2020)Chen, Tan, Ren, Xu, Sun, Zhao, Qin, and Liu}]{chen2020multispeech}
Mingjian Chen, Xu~Tan, Yi~Ren, Jin Xu, Hao Sun, Sheng Zhao, Tao Qin, and Tie-Yan Liu. 2020.
\newblock Multispeech: Multi-speaker text to speech with transformer.
\newblock \emph{arXiv preprint arXiv:2006.04664}.

\bibitem[{Ding et~al.(2021)Ding, Yang, Hong, Zheng, Zhou, Yin, Lin, Zou, Shao, Yang et~al.}]{ding2021cogview}
Ming Ding, Zhuoyi Yang, Wenyi Hong, Wendi Zheng, Chang Zhou, Da~Yin, Junyang Lin, Xu~Zou, Zhou Shao, Hongxia Yang, and 1 others. 2021.
\newblock Cogview: Mastering text-to-image generation via transformers.
\newblock \emph{Advances in neural information processing systems}, 34:19822--19835.

\bibitem[{Fujitake(2024)}]{fujitake2024dtrocr}
Masato Fujitake. 2024.
\newblock Dtrocr: Decoder-only transformer for optical character recognition.
\newblock In \emph{Proceedings of the IEEE/CVF winter conference on applications of computer vision}, pages 8025--8035.

\bibitem[{Graves(2013)}]{graves2013generating}
Alex Graves. 2013.
\newblock Generating sequences with recurrent neural networks.
\newblock \emph{arXiv preprint arXiv:1308.0850}.

\bibitem[{Haines et~al.(2016)Haines, Mac~Aodha, and Brostow}]{haines2016my}
Tom~SF Haines, Oisin Mac~Aodha, and Gabriel~J Brostow. 2016.
\newblock My text in your handwriting.
\newblock \emph{ACM Transactions on Graphics (TOG)}, 35(3):1--18.

\bibitem[{Kang et~al.(2020)Kang, Riba, Wang, Rusinol, Forn{\'e}s, and Villegas}]{kang2020ganwriting}
Lei Kang, Pau Riba, Yaxing Wang, Mar{\c{c}}al Rusinol, Alicia Forn{\'e}s, and Mauricio Villegas. 2020.
\newblock Ganwriting: content-conditioned generation of styled handwritten word images.
\newblock In \emph{Computer Vision--ECCV 2020: 16th European Conference, Glasgow, UK, August 23--28, 2020, Proceedings, Part XXIII 16}, pages 273--289. Springer.

\bibitem[{Li et~al.(2023)Li, Lv, Chen, Cui, Lu, Florencio, Zhang, Li, and Wei}]{li2023trocr}
Minghao Li, Tengchao Lv, Jingye Chen, Lei Cui, Yijuan Lu, Dinei Florencio, Cha Zhang, Zhoujun Li, and Furu Wei. 2023.
\newblock Trocr: Transformer-based optical character recognition with pre-trained models.
\newblock In \emph{Proceedings of the AAAI conference on artificial intelligence}, volume~37, pages 13094--13102.

\bibitem[{Li et~al.(2019)Li, Liu, Liu, Zhao, and Liu}]{li2019neural}
Naihan Li, Shujie Liu, Yanqing Liu, Sheng Zhao, and Ming Liu. 2019.
\newblock Neural speech synthesis with transformer network.
\newblock In \emph{Proceedings of the AAAI conference on artificial intelligence}, volume~33, pages 6706--6713.

\bibitem[{Liwicki and Bunke(2005)}]{liwicki2005iam}
Marcus Liwicki and Horst Bunke. 2005.
\newblock Iam-ondb-an on-line english sentence database acquired from handwritten text on a whiteboard.
\newblock In \emph{Eighth International Conference on Document Analysis and Recognition (ICDAR'05)}, pages 956--961. IEEE.

\bibitem[{Ma et~al.(2024)Ma, Wang, Jia, Chen, Liu, Li, Chen, and Qiao}]{ma2024latte}
Xin Ma, Yaohui Wang, Gengyun Jia, Xinyuan Chen, Ziwei Liu, Yuan-Fang Li, Cunjian Chen, and Yu~Qiao. 2024.
\newblock Latte: Latent diffusion transformer for video generation.
\newblock \emph{arXiv preprint arXiv:2401.03048}.

\bibitem[{Vaswani et~al.(2017)Vaswani, Shazeer, Parmar, Uszkoreit, Jones, Gomez, Kaiser, and Polosukhin}]{vaswani2017attention}
Ashish Vaswani, Noam Shazeer, Niki Parmar, Jakob Uszkoreit, Llion Jones, Aidan~N Gomez, {\L}ukasz Kaiser, and Illia Polosukhin. 2017.
\newblock Attention is all you need.
\newblock \emph{Advances in neural information processing systems}, 30.

\bibitem[{Yeleussinov et~al.(2023)Yeleussinov, Amirgaliyev, and Cherikbayeva}]{yeleussinov2023improving}
Arman Yeleussinov, Yedilkhan Amirgaliyev, and Lyailya Cherikbayeva. 2023.
\newblock Improving ocr accuracy for kazakh handwriting recognition using gan models.
\newblock \emph{Applied Sciences}, 13(9):5677.

\end{thebibliography}

\appendix
\newpage
\section{Appendix}
\label{sec:appendix}
\subsection{Training Configuration}
Our model is trained on 8 NVIDIA V100 GPUs with a per-GPU batch size of 64. We adopt the Adam optimizer with a learning rate of 0.0001. Both the encoder and decoder are implemented as 3-layer Transformers, each with 4 attention heads and a hidden dimension of 512. In the Gaussian memory mask, we set the scaling factor $r=17$ in Eq.~\ref{eq: center}. The mixture density network (MDN) outputs a 20-component Gaussian mixture ($K = 20$) to parameterize the pen trajectory distribution at each decoding step.

\subsection{Evaluation Pipelines}
\textbf{Subjective Evaluation}: We conducted a subjective evaluation with 20 human evaluators to score samples generated by four methods: AlexRNN and TrInk (both with and without the Top-k strategy). For the experiment, 96 text inputs were used to generate samples, and each output was rated on two criteria—style consistency and legibility using a 5-point Likert scale (1: lowest, 5: highest). Higher scores indicate better performance.

\noindent \textbf{Objective Evaluation}: We first convert the generated handwriting samples into standardized textline images to simulate realistic OCR application scenarios. These images are then fed into the state-of-the-art TrOCR model \cite{li2023trocr} for text recognition. The outputs from TrOCR are systematically compared with the ground-truth text to compute Character Error Rate (CER) and Word Error Rate (WER), which quantify character-level inaccuracies and word-level mismatches, respectively. Notably, WER is excluded for short-text evaluations due to its instability when applied to limited word counts, as minor errors disproportionately skew the metric. Lower CER values indicate higher legibility, providing an automated and reproducible measure of text quality. During the evaluation of style equalization, we dynamically sample style inputs from the training set for the style encoder, ensuring that each synthesized handwriting sample corresponds to a unique, randomly selected style reference from the training dataset. 

\begin{figure}[!t]
    \centering
    \includegraphics[width=\linewidth]{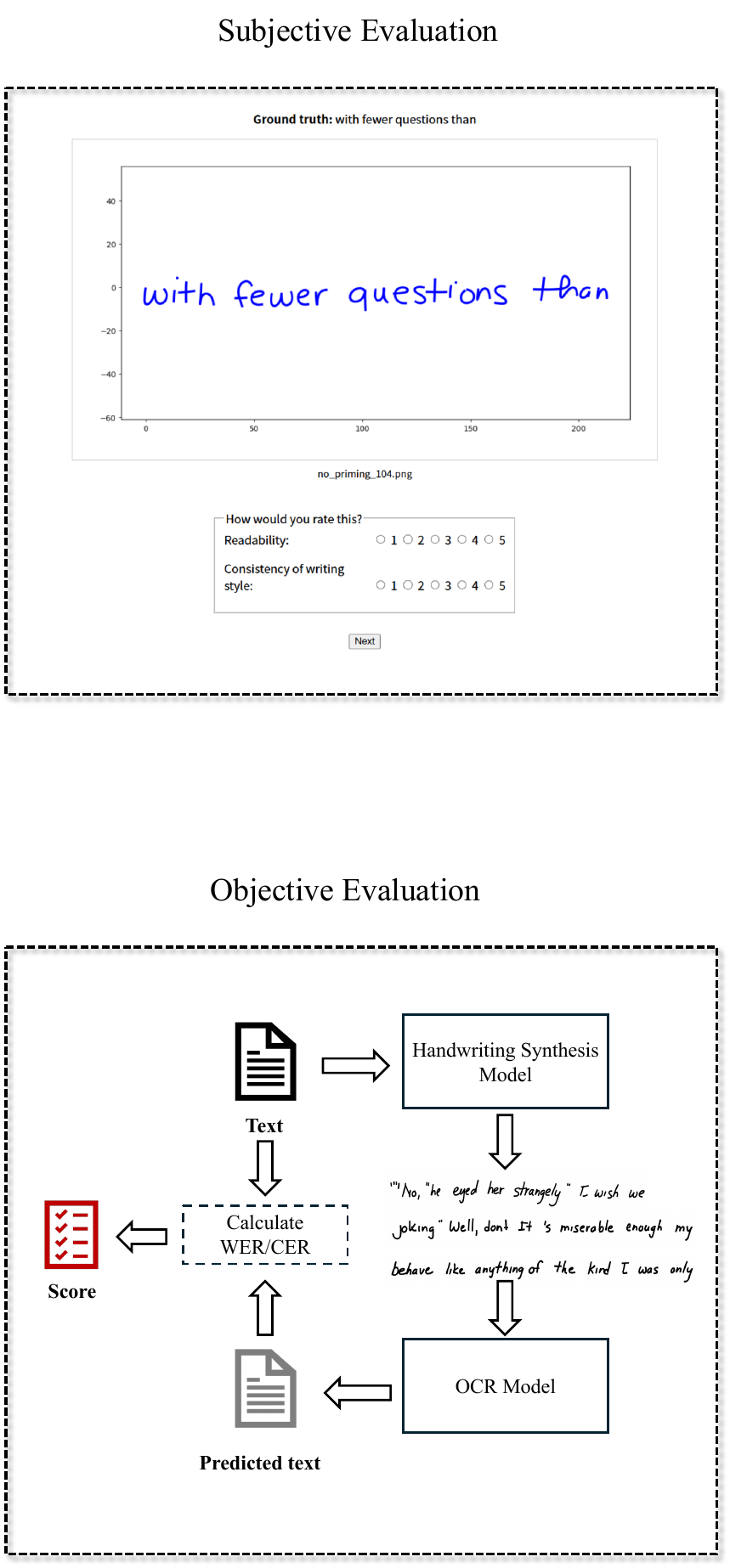}
    \caption{Subjective and Objective Evaluation Pipelines.}
    \label{fig:eval}
\end{figure}
\subsection{Visualization of Generated Samples}
We present a collection of generated handwriting samples based on 13 text prompts of varying lengths. Each row illustrates outputs from four models: AlexRNN, AlexRNN (Top-k), TrInk, and TrInk (Top-k), displayed from left to right. As observed, TrInk consistently produces handwriting that is more legible and stylistically consistent than that of the RNN-based.
\begin{figure*}[!t]
    \centering
    \includegraphics[width=\linewidth]{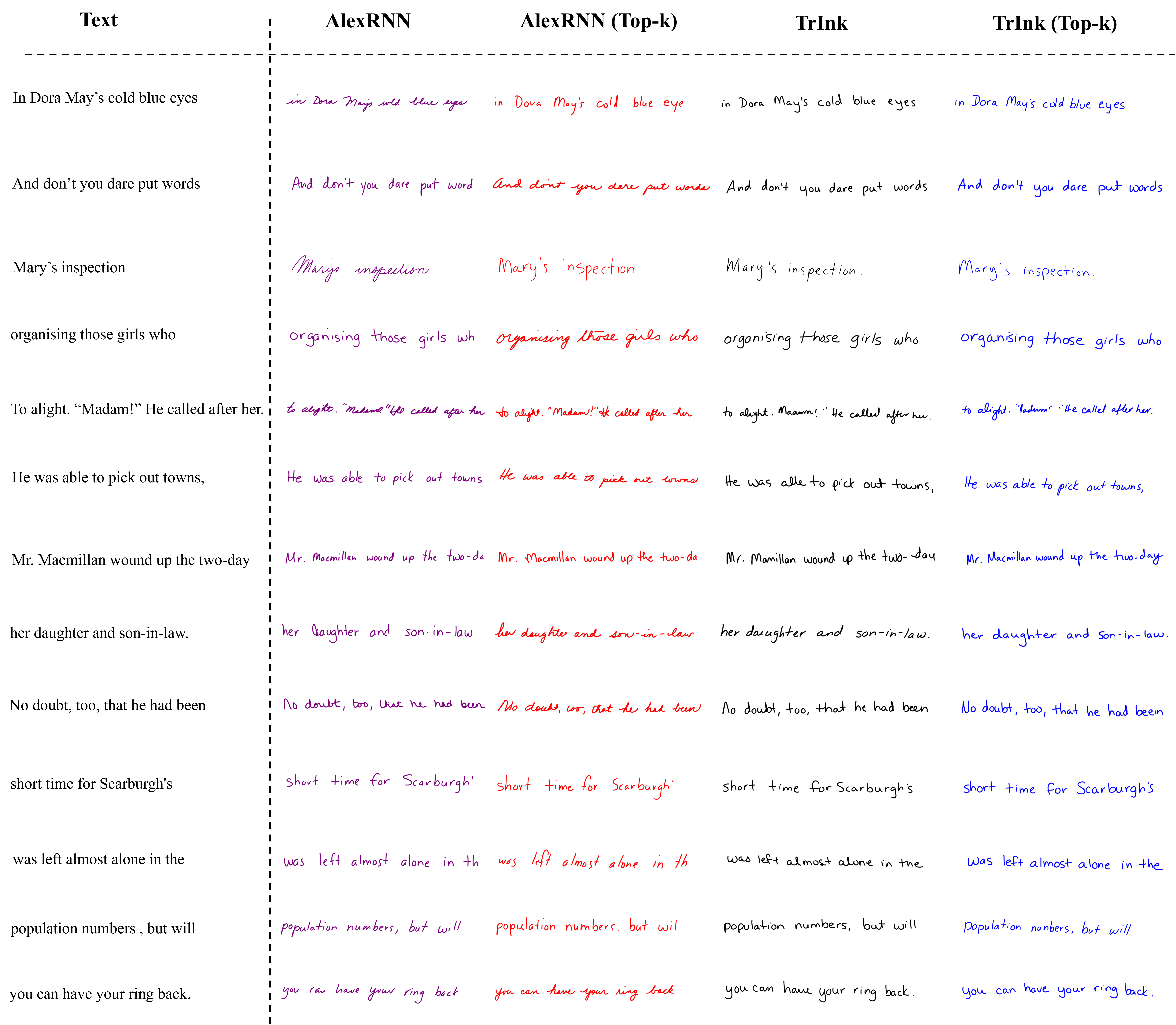}
    \caption{Samples for AlexRNN and TrInk.}
    \label{fig:enter-label}
\end{figure*}
\end{document}